\documentclass[runningheads]{llncs}
\usepackage[T1]{fontenc}
\usepackage{graphicx,verbatim}
\usepackage{multirow}
\usepackage{amsmath}
\usepackage{stfloats}
\usepackage{bm}
\usepackage{makecell}
\usepackage{color}
\usepackage{booktabs}
\usepackage{amssymb}
\usepackage[table]{xcolor}
\usepackage{xcolor}
\usepackage{marvosym}

\begin{document}
%
\title{Towards Clinically Interpretable Ophthalmic VQA via Spatially-Grounded Lesion Evidence}
\titlerunning{Lesion-Aware Ophthalmic VQA}
%

\author{Xingyue Wang\inst{1} \and
Bo Liu\inst{2} \and
Meng Wang\inst{3} \and
Zhixuan Zhang\inst{1} \and
Chengcheng Zhu\inst{4} \and
Huazhu Fu\inst{5}\textsuperscript{\Letter} \and
Jiang Liu\inst{1}\textsuperscript{\Letter}}
\authorrunning{X. Wang et al.}
%
\institute{Department of Computer Science and Engineering, Southern University of Science and Technology, Shenzhen 518055, China
\and The Hong Kong Polytechnic University, Hong Kong S.A.R., China.
\and National University of Singapore, Singapore.
\and University of Washington,  Seattle, WA, USA.
\and Institute of High Performance Computing, Agency for Science, Technology and Research, Singapore. \\
\email{liuj@sustech.edu.cn, hzfu@ieee.org}
}

  
\maketitle              
\begin{abstract}

Visual Question Answering (VQA) holds great promise for clinical support, particularly in ophthalmology, where retinal fundus photography is essential for diagnosis. 
However, existing ophthalmic VQA benchmarks primarily emphasize answer accuracy, neglecting the explicit visual evidence necessary for clinical interpretability.
In this work, we introduce \textbf{ FundusGround}, a new benchmark for clinically interpretable ophthalmic VQA with spatially-grounded lesion evidence.
Specifically, we propose a three-stage pipeline that collects 10,719 fundus images with 15,595 image-level meticulously annotated lesions. To ensure anatomical consistency and clinical validity, all lesions are spatially localized using the Early Treatment Diabetic Retinopathy Study (ETDRS) grid, enabling standardized mapping to clinically meaningful retinal regions.
Built upon this structured lesion evidence, 72,706 questions are then generated spanning four formats: open-ended, closed-ended, single-choice, and multiple-choice.
We further benchmark multiple general- and medical- large vision-language models using dual metrics for answer accuracy and lesion-level reasoning. The experiments demonstrate that incorporating lesion-level visual evidence consistently improves model performance and transparency, highlighting the necessity of explicit spatial grounding for reliable and explainable ophthalmic VQA. The project page is \url{https://www.med-vqa.com/FundusGround}.


\end{abstract}
\section{Introduction}



Medical Visual Question Answering (VQA)~\cite{liu2021slake} enables clinical decision support by generating natural language answers to specific medical queries through joint reasoning over medical images and questions. The recent evolution of large vision-language models (LVLMs) has significantly enhanced performance across diverse medical benchmarks, particularly in radiology and pathology~\cite{chen2024wsi,liu2025gemex,Liu_2025_ICCV}. However, extending LVLMs to ophthalmology is particularly challenging, as retinal diagnosis demands fine-grained, lesion-level, and spatially-aware reasoning.


In ophthalmology, fundus photography is the gold standard for screening and managing vision-threatening conditions~\cite{grzybowski2024retina} such as diabetic retinopathy (DR) and age-related macular degeneration (AMD). 
Effective retinal assessment requires the identification of fine-grained lesions and their precise localization relative to anatomical landmarks~\cite{wang2021deep,kamble2022laden,chen2025mimo}.
Importantly, clinical grading and treatment decisions further depend on whether lesions occur within clinically significant retinal zones, which can be systematically referenced using frameworks such as the Early Treatment Diabetic Retinopathy Study (ETDRS) grid~\cite{attiku2023comparison}.

Despite its essential role in clinical decision-making, lesion-level localization remains insufficiently evaluated in existing ophthalmic VQA benchmarks~\cite{wei2025funbench,xu2025benchmarking,li2025eyecaregpt}.
Most datasets prioritize final answer correctness while not providing the explicit visual evidence required to justify clinical decisions. This ``black-box'' limits the interpretability and reliability of models in high-stakes medical scenarios. 
Furthermore, retinal pathologies are inherently heterogeneous and spatially complex, yet current benchmarks often fail to model the co-occurrence of multiple lesion types across different retinal subregions.
In addition, the restriction to a single question format in many datasets prevents a comprehensive evaluation of a model's reasoning capabilities across diverse clinical queries.

To address these limitations, we introduce \textbf{FundusGround}, the first lesion-aware multi-type ophthalmic VQA benchmark designed to bridge the gap between model prediction and clinically verifiable evidence. Our contributions are:
\begin{itemize}
    \item \textbf{Clinically-Aligned Spatial Mapping:} We introduce a lesion-to-region mapping framework that standardizes 15,595 retinal lesions from diverse fundus images onto a standardized ETDRS grid. This design enables consistent, spatially-grounded reasoning aligned with clinical practice.

    \item \textbf{Evidence-Anchored Multi-Type VQA Construction:} Built upon the ETDRS-grounded lesion evidence, we construct FundusGround with four question formats (open-ended, close-ended, single-choice, and multi-choice).  
    Each answer is explicitly linked to specific lesion types and ETDRS regions, facilitating fine-grained visual explanations.

    \item \textbf{Interpretability-Focused Evaluation:} We propose a dual-metric evaluation protocol that quantifies both answer accuracy and the alignment of model reasoning with ground-truth lesion type and location.
    
    \item \textbf{Performance Enhancement via Evidence:} Comprehensive experiments demonstrate that supervised fine-tuning with lesion-level evidence consistently improves answer correctness and provides fine-grained visual explanations, particularly for models with limited capacity.
    
\end{itemize}


\section{Construction of FundusGround}

\begin{figure}[!t]
\centering
\includegraphics[width=\textwidth]{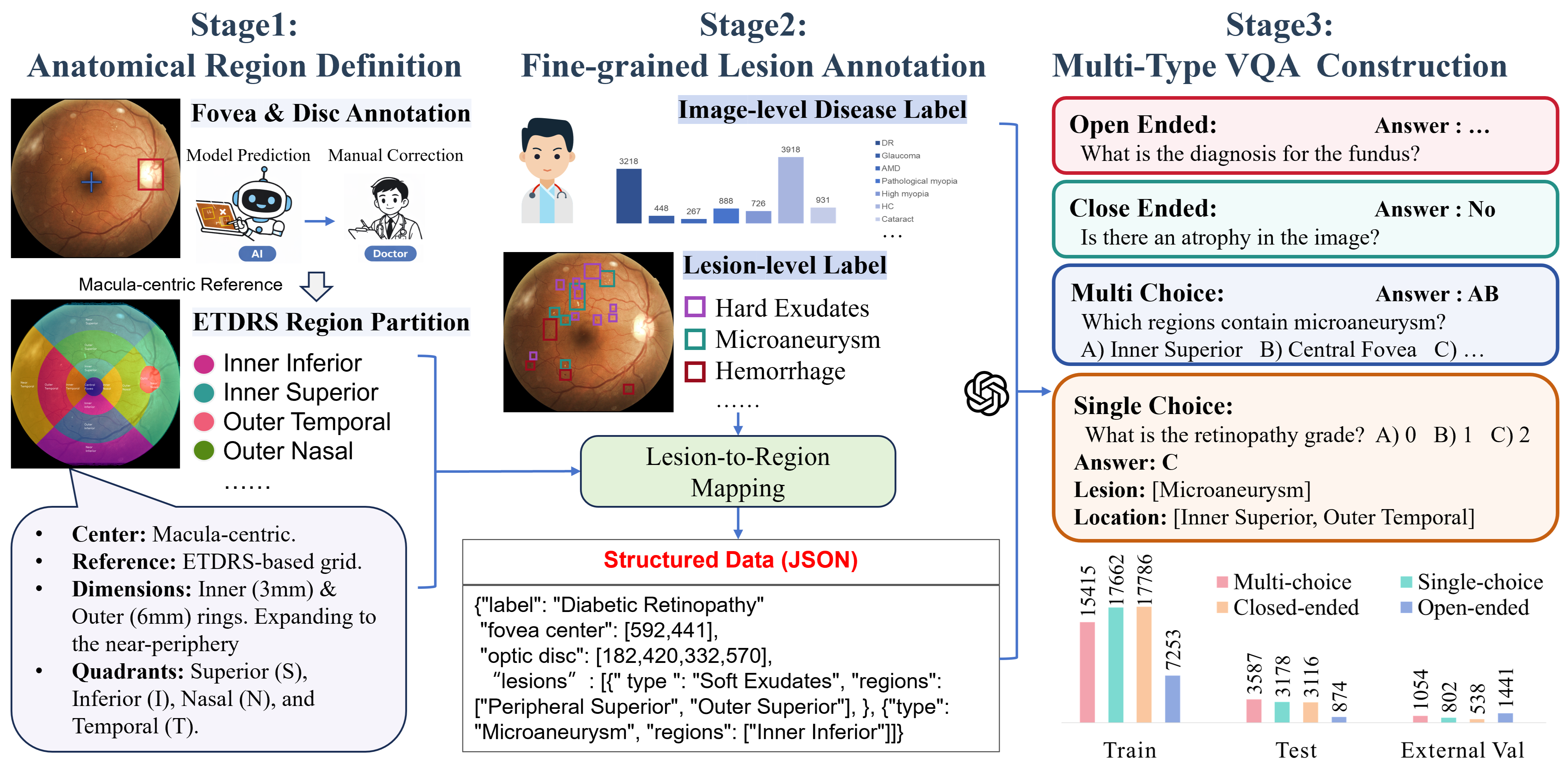}
\caption{Three-stage pipeline for constructing FundusGround.
Stage 1 defines clinically meaningful retinal regions using the ETDRS grid;
Stage 2 outputs fine-grained visual annotations (in both image- and lesion- levels);
Stage 3 constructs and filters multiple ophthalmic VQA types grounded in the structured lesion-aware visual evidence.} \label{framework}
\end{figure}

We propose a three-stage pipeline for constructing FundusGround, as illustrated in Fig. \ref{framework}. Our dataset focuses on fundus images with detailed lesion annotations, integrating lesion type and anatomical location to support multiple VQA tasks.

\subsection{Anatomical Region Definition}

To standardize lesion localization across heterogeneous fundus images, we first collect and unify data from 10 publicly available ophthalmic datasets, covering a wide range of retinal diseases. Specifically, these datasets include diabetic retinopathy (IDRID \cite{porwal2018indian}, Retinal-Lesions \cite{wei2021learn}, FGADR \cite{zhou2020benchmark}), cataract \footnote{https://www.kaggle.com/datasets/drskprabhakar/cataract-dr-normal-glaucoma-fundus-images-dataset/data}, pathological myopia (PALM \cite{palm2019fu} and additional in-house datasets), high myopia (HPMI \cite{huang2023hpmi}), age-related macular degeneration (ADAM \cite{fang2022adam}), and glaucoma-related fundus datasets (ORIGA \cite{zhang2010origa}, REFUGE \cite{orlando2020refuge}). In total, 10,719 fundus images are included, comprising 4,045 normal images, 3,266 diabetic retinopathy (DR), 458 glaucoma, 225 age-related macular degeneration (AMD), 841 cataract, 968 high myopia, and 916 pathological myopia cases, as detailed in Fig.~\ref{data_statistic}(a). 

To enable consistent and clinically meaningful spatial reasoning, we adopt the Early Treatment Diabetic Retinopathy Study (ETDRS) grid \cite{attiku2023comparison} as the reference framework (Stage 1 in Fig.~\ref{framework}). For each fundus image, we first localize the foveal center and the optic disc to establish a macula-centered coordinate system. When annotations are available in the original datasets, we directly use the provided labels. For images without such annotations, we adopt a detection model trained on the annotated subset to predict the optic disc and fovea locations. The predicted landmarks are then manually reviewed and corrected to ensure spatial accuracy and clinical reliability. Based on the macular center, the retina is partitioned into ETDRS regions, including the central macula, inner ring, and outer ring. The inner and outer rings are further divided into four quadrants: superior (S), inferior (I), nasal (N), and temporal (T). This macula-centered framework can be extended toward the near-peripheral retina when broader spatial coverage is required.


\subsection{Fine-grained Lesion Annotation}

\begin{figure}[!t]
\centering
\includegraphics[width=\textwidth]{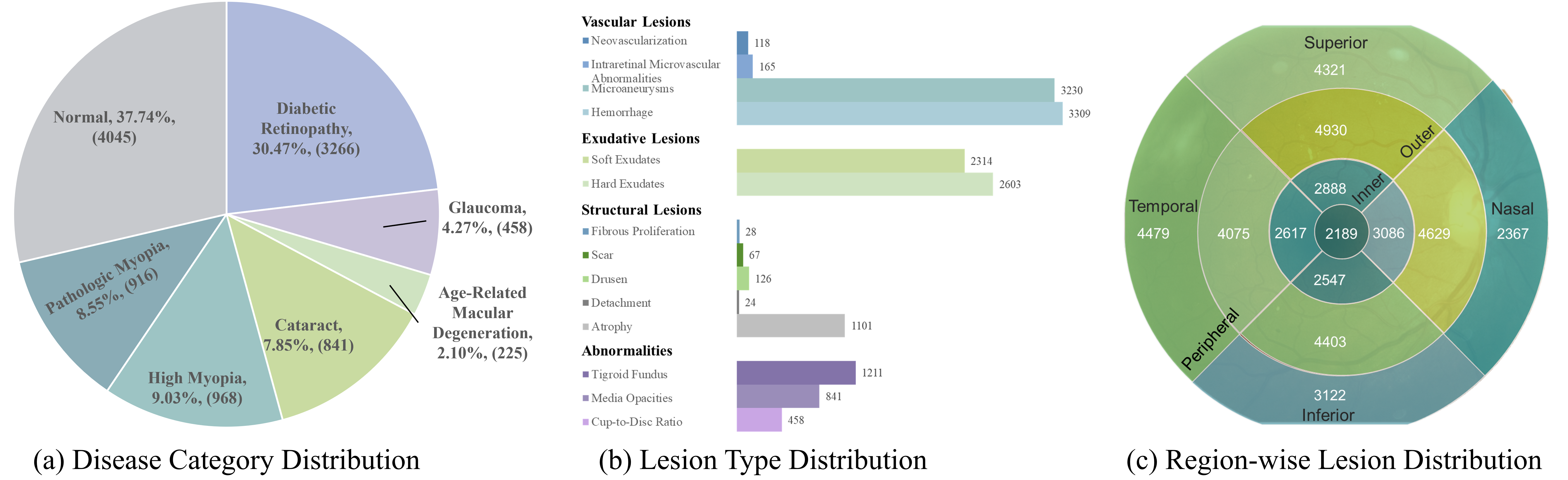}
\caption{Dataset statistics of FundusGround, including disease category distribution (a), lesion type distribution (b), and region-wise lesion distribution (c).} \label{data_statistic}
\end{figure}

Building upon the standardized anatomical regions established in Stage 1, we perform fine-grained lesion annotation by leveraging the lesion labels provided in the original datasets. As a result, each fundus image is enriched with structured annotations at multiple levels, including disease-level labels (e.g., DR) and lesion-level evidence (e.g., \emph{lesion type} and corresponding \emph{ETDRS region}), as illustrated in the structured data example in Stage 2 of Fig.~\ref{framework}. 

Specifically, we map each lesion to the ETDRS regions established in Stage 1. For each image, the macular center and optic disc are first localized, forming a macula-centric coordinate system. Then, lesion location $l_i$, represented by a bounding box (or segmentation mask) in the original datasets, is then projected onto the ETDRS grid $R_j$, resulting in a lesion-to-region mapping, defined as:
\begin{equation}
    M(l_i) = \{R_j \mid \text{Area}(l_i \cap R_j) / \text{Area}(l_i) > \tau \},
\end{equation}
where $\tau$ is the overlapping thresholding (set 0 in this work).
When a lesion spans multiple ETDRS subregions, all corresponding regions are recorded to capture its spatial extent faithfully.
Note that ophthalmologists review the lesion mappings to resolve ambiguities and ensure clinical consistency, particularly in cases involving overlapping lesions or borderline regions. 


In total, 15,595 retinal lesions are curated across diverse types, as illustrated in Fig.~\ref{data_statistic}(b): 6,822 vascular lesions (e.g., microaneurysms and hemorrhages), 4,917 exudative lesions (e.g., hard and soft exudates), 1,346 structural lesions (e.g., drusen and retinal atrophy), and 2,510 other abnormalities. This distribution reflects the heterogeneity, co-occurrence, and spatial complexity observed in real-world ophthalmic practice.
After lesion-to-region mapping, we quantify the regional distribution of lesions on the ETDRS grid. Fig.~\ref{data_statistic}(c) demonstrates that lesions are distributed across all retinal regions without pronounced regional imbalance, ensuring comprehensive spatial coverage. Note that lesion-region relationships are not one-to-one: multiple lesion types may co-exist within a region, and individual lesions may extend across regional boundaries.

Finally, all lesion information is organized into a structured, machine-readable format, explicitly linking disease labels, anatomical landmarks, and lesion-level visual evidence in a unified representation. This structured representation ensures that every lesion is explicitly associated with both a semantic type and a precise retinal subregion, providing the spatially grounded lesion evidence required for clinically interpretable and reliable ophthalmic VQA.

\subsection{Comprehensive Multi-type VQA Construction}

Based on the fine-grained lesion annotations, we construct a comprehensive ophthalmic VQA benchmark encompassing open-ended, closed-ended, single-choice, and multi-choice question formats, as in Fig.~\ref{framework}(stage 3). 
All questions are explicitly designed not only to require correct answers, but also to necessitate the identification of the corresponding lesion types and their anatomical regions as supporting evidence.
An example that targets overall disease status can be ``Does this fundus image show signs of diabetic retinopathy? Answer: Yes, Lesion: Soft Exudates, Region: Outer Superior''. 

The question generation follows a human–AI collaborative pipeline. Candidate questions are first generated automatically using GPT-5~\cite{singh2025openai}, conditioned on the structured lesion annotations and ETDRS regions. These automatically generated questions are then filtered through a combination of rule-based checks and Qwen3.5\cite{qwen3.5}-assisted screening to remove unanswerable, ambiguous, or clinically implausible queries. Subsequently, ophthalmologists review a subset of questions to verify clinical correctness, clarity, and relevance, and their feedback is used to refine the filtering process iteratively. 


After filtering, the resulting dataset contains a total of 15,415 multi-choice questions, 17,662 single-choice questions, 17,786 closed-ended questions, and 7,253 open-ended questions in the training split, with 3,587, 3,178, 3,116, and 874 questions in the test split, respectively. To evaluate generalization, an external validation set is constructed using two entirely independent ophthalmic datasets, contributing 1,054, 802, 538, and 1,441 questions across multi-choice, single-choice, closed-ended, and open-ended formats, respectively. No images or questions from these external datasets are used during training. This construction balances question diversity, lesion coverage, and task complexity, providing a unified, lesion-aware benchmark for systematic evaluation of model performance across heterogeneous ophthalmic VQA scenarios. The dataset, along with the prompts used to generate the data, will be open-sourced upon acceptance.

\section{Experiments}

\subsection{Evaluation Setup}

We evaluate a range of representative large vision–language models, including general models (GPT-4o \cite{achiam2023gpt}, InternVL3 \cite{zhu2025internvl3}, Qwen3-VL \cite{bai2025qwen3}) and medical ones (Lingshu  \cite{xu2025lingshu}, Medgemma  \cite{sellergren2025medgemma}, Medgemma-1.5  \cite{sellergren2025medgemma}) to systematically benchmark performance on FundusGround. For GPT-4o, 100 cases per task are randomly sampled due to API constraints. The evaluation is conducted across three distinct settings:
(1) \textbf{Zero-shot}: Models are prompted to output both the answer and lesion-level evidence, with regular expressions used to extract structured information for evaluation; 
(2) Supervised fine-tuning (SFT) without lesion supervision (\textbf{SFT w/o L}): Models are fine-tuned to generate answer-only outputs in the format: \texttt{<response>}
\texttt{<answer>...</answer>} 
\texttt{</response>}; 
(3) SFT with lesion-level supervision (\textbf{SFT w/ L}) introduces explicit lesion-level supervision, requiring models to produce both structured lesion evidence and the final answer in \texttt{<response>} 
\texttt{<lesion>} 
\texttt{<type>...</type>} 
\texttt{<location>...</location>} 
\texttt{</lesion>} 
\texttt{<answer>...</answer>} 
\texttt{</response>}. 
This structured format enables reliable parsing and separate evaluation of answer correctness and lesion-level reasoning, regardless of the underlying model architecture.






\subsection{Evaluation Metrics}




Model performance on FundusGround is evaluated along two dimensions: answer correctness and lesion-level reasoning, allowing assessment of both clinical accuracy and reasoning grounded in visual evidence.

\textbf{Answer Score}. Correctness is evaluated according to question type: \emph{accuracy} is used for multiple-choice, single-choice, and closed-ended questions, while \emph{BLEU-1} is adopted for open-ended, free-text responses. This evaluation scheme ensures standardized and fair assessment across diverse question formats.

\textbf{Lesion Evidence Score}. Lesion-level reasoning is assessed with two metrics: Lesion Type F1 (how accurately lesion categories are identified) and Lesion Location F1 (agreement with ground-truth ETDRS regions). Together, they indicate how well the model links answers to anatomical and pathological evidence, providing a clinically interpretable measure of reasoning.

\subsection{Results and Analysis}

\begin{table}[t]
\centering
\caption{Performance of representative LVLMs on the FundusGround test set across four question types. 
Evaluation settings include Zero-shot (no fine-tuning), SFT w/o L (finetuned using answer supervision only), and SFT w/ L (finetuned with both answer and lesion-level supervision). 
Metrics (in \%): A denotes answer correctness; T denotes lesion type F1 score; L denotes lesion location F1 score. }
\setlength{\tabcolsep}{1.5pt} 
\scriptsize
\begin{tabular}{l l ccc ccc ccc ccc}
\toprule
\multirow{2}{*}{Model} & \multirow{2}{*}{Setting} 
& \multicolumn{3}{c}{Open-ended} 
& \multicolumn{3}{c}{Close-ended} 
& \multicolumn{3}{c}{Single-choice} 
& \multicolumn{3}{c}{Multi-choice} \\
\cmidrule(lr){3-5} \cmidrule(lr){6-8} \cmidrule(lr){9-11} \cmidrule(lr){12-14}
& & A & T & L & A & T & L & A & T & L & A & T & L \\
\midrule
\makecell[l]{GPT-4o} & Zero-shot & 13.01 & 5.32 & 0.49 & 45.00 & 4.56 & 1.26 & 40.00 & 5.05 & 0.51 & 7.00 & 7.72 & 3.89 \\
\midrule
\multirow{3}{*}{\makecell[l]{Qwen3-VL\\-8B}} 
 & Zero-shot &14.70 & 11.09 & 6.01 & 26.52 & 9.85 & 3.21 & 49.52 & 8.98 & 5.21 & 19.52 & 12.36 & 2.68 \\
 & SFT w/o L & 32.24 & - & - & 45.81 & - & - & 78.24 & - & - & 62.34 & - & - \\
 & \multicolumn{13}{>{\columncolor{gray!15!white}}c}{} \\[-2.2ex] 
 & SFT w/ L & 36.49 & 40.48 & 24.29 & 54.22 & 29.19 & 21.03 & 85.64 & 41.96 & 32.04 & 69.32 & 34.70 & 23.37 \\
\midrule
\multirow{3}{*}{\makecell[l]{InternVL3\\-8B}} 
 & Zero-shot & 9.42 & 1.94 & 0.26 & 40.06 & 1.03 & 0.28 & 57.35 & 2.67 & 1.01 & 17.89 & 1.37 & 0.61 \\
 & SFT w/o L & 21.82 & - & - & 48.98 & - & - & 70.58 & - & - & 25.43 & - & - \\
 & \multicolumn{13}{>{\columncolor{gray!15!white}}c}{} \\[-2.2ex] 
 & SFT w/ L & 38.42 & 41.52 & 24.16 & 52.11 & 40.48 & 29.82 & 82.36 & 48.92 & 36.82 & 69.52 & 41.26 &  34.52 \\
\midrule
\multirow{3}{*}{\makecell[l]{Lingshu-7B}} 
 & Zero-shot & 23.07 & - & - & 47.22 & - & - & 51.09 & - & - & 20.88 & - & - \\
 & SFT w/o L & 27.49 & - & - & 51.28 & - & - & 74.14 & - & - & 40.12 & - & - \\
 & \multicolumn{13}{>{\columncolor{gray!15!white}}c}{} \\[-2.2ex] 
 & SFT w/ L & 30.29 & 35.18 & 29.68 & 55.13 & 38.48 & 22.68 & 85.36 & 44.13 & 33.21 & 57.25 & 38.12 & 28.14 \\
\midrule
\multirow{3}{*}
{\makecell[l]{Medgemma\\-4B}} 
 & Zero-shot & 11.11 & 0.90 & 0.15 & 28.24 & 0.5 & 0.16 & 57.19 & 0.63 & 0.24 & 12.35 & 0.57 & 0.22 \\
 & SFT w/o L & 17.45 & - & - & 39.27 & - & - & 70.13 & - & - & 30.02 & - & - \\
 & \multicolumn{13}{>{\columncolor{gray!15!white}}c}{} \\[-2.2ex] 
 & SFT w/ L & 30.87 & 37.73 & 23.96 & 51.60 & 45.39 & 32.07 & 84.17 & 49.82 & 39.41 & 64.01 & 43.81 & 31.07 \\
\midrule
\multirow{3}{*}{\makecell[l]{Medgemma\\1.5-4B}} 
 & Zero-shot & 7.87 & 2.98 & 0.94 & 31.65 & 1.63 & 0.72 & 19.48 & 1.95 & 0.75 & 20.38 & 1.39 & 0.65 \\
 & SFT w/o L & 18.27 & - & - & 42.63 & - & - & 68.91 & - & - & 43.77 & - & - \\
 & \multicolumn{13}{>{\columncolor{gray!15!white}}c}{} \\[-2.2ex] 
 & SFT w/ L & 35.24 & 48.27 & 36.29 & 54.22 & 44.18 & 33.64 & 88.60 & 50.00 & 40.17 & 70.24 & 43.49 & 33.11 \\
\bottomrule
\end{tabular}
\label{test}
\end{table}

We present the results on the test set and the external validation set in Table \ref{test} and Table \ref{val_external}, respectively, for a detailed analysis.

\textbf{Overall Performance.} First, zero-shot performance is limited, especially for lesion-aware reasoning.
Across both test and external validation sets, all models exhibit low Answer-scores on open-ended and multi-choice tasks, along with consistently poor lesion-type and lesion-location performance. Even strong general models such as GPT-4o and Qwen3-VL fail to reliably identify and localize retinal lesions, indicating a lack of explicit lesion-aware reasoning capabilities.

\begin{table}[t]
\centering
\caption{Performance of LVLMs on the external validation set of FundusGround.}
\setlength{\tabcolsep}{1.5pt} 
\scriptsize
\begin{tabular}{l l ccc ccc ccc ccc}
\toprule
\multirow{2}{*}{Model} & \multirow{2}{*}{Setting} 
& \multicolumn{3}{c}{Open-ended} 
& \multicolumn{3}{c}{Close-ended} 
& \multicolumn{3}{c}{Single-choice} 
& \multicolumn{3}{c}{Multi-choice} \\
\cmidrule(lr){3-5} \cmidrule(lr){6-8} \cmidrule(lr){9-11} \cmidrule(lr){12-14}
& & A & T & L & A & T & L & A & T & L & A & T & L \\
\midrule
\makecell[l]{GPT-4o} & Zero-shot & 15.25 & 5.66 & 1.01 & 44.00 & 6.44 & 0.22 & 47.00 & 9.92 & 1.24 & 23.00 & 2.76 & 0.15 \\
\midrule
\multirow{3}{*}{\makecell[l]{Qwen3-VL\\-8B}} 
 & Zero-shot & 12.48 & 9.52 & 2.28 & 28.44 & 6.74 & 2.69 & 52.29 & 11.46 & 4.97 & 21.56 & 12.78 & 4.92 \\
 & SFT w/o L & 22.57  & - & - & 56.89 & - & - & 82.86 & - & - & 53.37 & - & - \\
 & \multicolumn{13}{>{\columncolor{gray!15!white}}c}{} \\[-2.2ex] 
 & SFT w/ L & 32.17 & 20.67 & 13.66 & 61.91 & 13.98 & 9.07 & 85.58 & 18.21 & 12.57 & 61.12 & 19.83 & 13.00\\
\midrule
\multirow{3}{*}{\makecell[l]{InternVL3\\-8B}} 
 & Zero-shot & 16.22 & 6.66 & 0.31 & 47.92 & 1.94 & 0.21 & 61.16 & 6.47 & 0.93 & 26.69 & 0.65 & 0.08 \\
 & SFT w/o L & 23.75 & - & - & 57.49 & - & - & 80.28 & - & - & 51.76 & - & - \\
 & \multicolumn{13}{>{\columncolor{gray!15!white}}c}{} \\[-2.2ex] 
 & SFT w/ L & 32.48 & 25.14 & 15.41 & 61.42 & 21.48 & 13.48 & 87.51 & 30.48 & 18.49 & 57.49 & 27.11 & 10.08 \\
\midrule
\multirow{3}{*}{\makecell[l]{Lingshu-7B}} 
 & Zero-shot & 22.34 & - & - & 51.58 & - & - & 52.52 & - & - & 19.66 & - & - \\
 & SFT w/o L & 26.51 & - & - & 54.83 & - & - & 79.18 & - & - & 50.24 & - & - \\
 & \multicolumn{13}{>{\columncolor{gray!15!white}}c}{} \\[-2.2ex] 
 & SFT w/ L & 27.13 & 19.48 & 8.79 & 56.14 & 21.49 & 12.47 & 87.58 & 26.94 & 14.63 & 52.19 & 29.74 & 13.48 \\
\midrule
\multirow{3}{*}{\makecell[l]{Medgemma\\-4B}} 
 & Zero-shot & 24.20 & 5.44 & 6.59 & 35.17 & 5.06 & 1.38 & 60.44 & 11.11 & 4.03 & 19.10 & 3.21 & 0.55 \\
 & SFT w/o L & 26.75 & - & - & 49.29 & - & - & 61.76 & - & - & 32.30 & - & - \\
 & \multicolumn{13}{>{\columncolor{gray!15!white}}c}{} \\[-2.2ex] 
 & SFT w/ L & 35.32 & 23.50 & 19.64 & 68.85 & 24.49 & 18.52 & 91.75 & 30.39 & 15.25 & 62.64 & 25.11 & 18.89 \\
\midrule
\multirow{3}{*}{\makecell[l]{Medgemma\\1.5-4B}} 
 & Zero-shot & 16.06 & 1.34 & 0.65 & 45.52 & 0.78 & 0.20 & 59.87 & 1.29 & 0.37 & 10.67 & 0.80 & 0.10 \\
 & SFT w/o L & 22.67 & - & - & 58.03 & - & - & 78.48 & - & - & 49.18 & - & - \\
 & \multicolumn{13}{>{\columncolor{gray!15!white}}c}{} \\[-2.2ex] 
 & SFT w/ L & 31.49 & 25.58 & 16.85 & 65.03 & 17.08 & 12.54 & 90.89 & 43.64 & 27.79 & 60.67 & 28.71 & 19.53 \\
\bottomrule
\end{tabular}
\label{val_external}
\end{table}

Next, supervised fine-tuning improves answer correctness but does not ensure lesion-level interpretability.
After SFT without lesion supervision, models achieve substantial gains in overall Answer-score across all tasks. 
For example, after fine-tuning, Qwen3-VL-8B achieves nearly a 42\% improvement on the multi-choice task, while MedGemma shows an improvement of nearly 49\% on the single-choice task in the test set.
However, since lesion-level learning is not incorporated during the SFT process, the associated abilities for lesion type discrimination and localization are largely forgotten~\cite{kaushik2021understanding}.

Finally, incorporating lesion-level visual evidence consistently yields the best performance.
In the SFT w/ L setting, all models are required to first generate lesion-level evidence before producing the final answer. This training paradigm, reasoning over lesions prior to answer prediction, not only impressively improves answer accuracy (compared to zero-shot and SFT w/o L settings), but also yields explicit anatomical localization.
For example, on the multi-choice task, MedGemma-1.5-4B achieves lesion type and location scores of \textbf{43.49\%} and \textbf{33.11\%}, respectively, while improving answer accuracy by \textbf{26.47\%} compared to SFT w/o L. (Table \ref{test}). These gains are consistent across all question types and both test and external validation sets, demonstrating that lesion-level supervision enables clinically-grounded and interpretable ophthalmic reasoning.

\begin{figure}[!t]
\centering
\includegraphics[width=\textwidth]{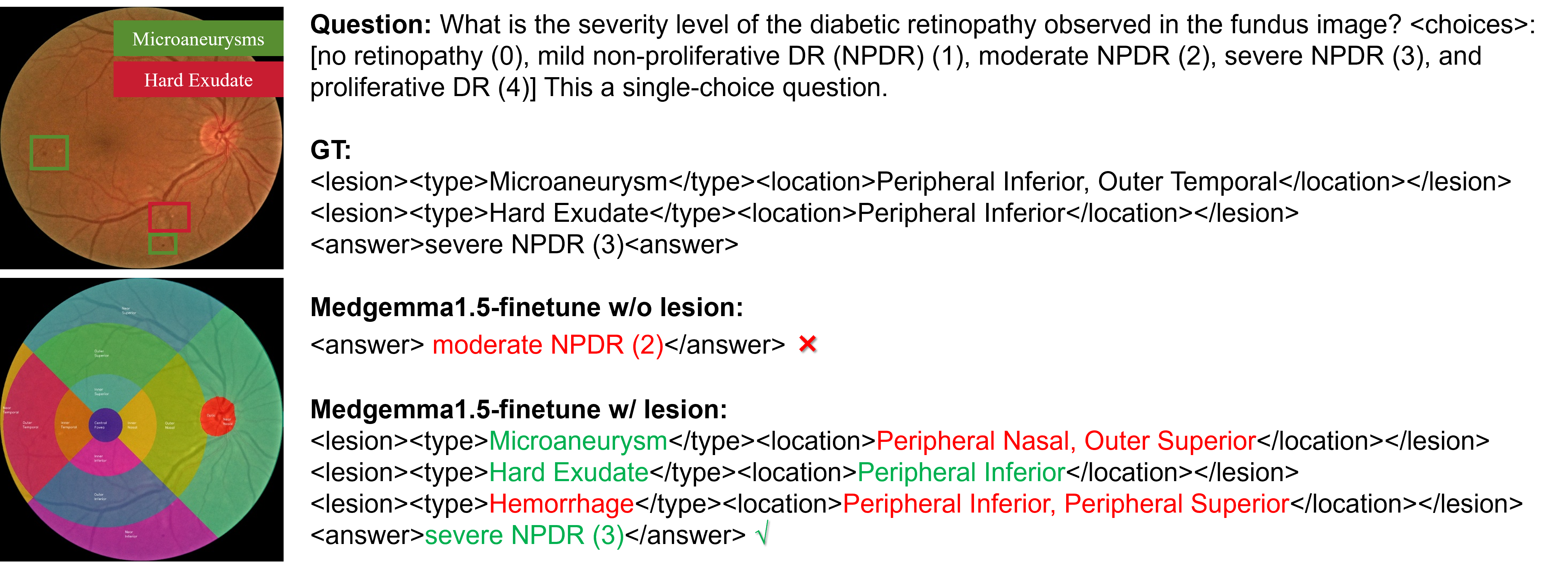}
\caption{Qualitative comparison on a DR grading example. \textcolor{green}{Green} denotes correct predictions, while \textcolor{red}{red} denotes incorrect parts.} \label{case1}
\end{figure}

\textbf{Case Study.} As shown in Fig. \ref{case1}, this example illustrates the impact of lesion-level supervision on diagnostic accuracy and interpretability. Without lesion supervision, MedGemma-1.5 underestimates disease severity by predicting moderate NPDR (2) and provides no explicit lesion evidence. In contrast, the lesion-aware model (SFT w/ L) correctly predicts severe NPDR (3) and identifies multiple clinically relevant lesions, consistent with the pathological characteristics of advanced disease.
Although lesion-level supervision improves pathological reasoning, predicted lesion locations do not fully align with ground truth, indicating that precise anatomical grounding remains challenging. Overall, this case demonstrates that FundusGround differentiates superficial answer correctness from true lesion-aware reasoning, and that explicit lesion type and location supervision are essential for clinically interpretable and accurate predictions.

\section{Conclusion}

We present FundusGround, a novel lesion-aware ophthalmic VQA benchmark that integrates fine-grained lesion annotations with spatially-grounded visual evidence, thereby enhancing the interpretability of model reasoning. 
The benchmark includes four distinct question types (i.e., open-ended, closed-ended, single-choice, and multi-choice), designed to evaluate both answer correctness and the model’s ability to reason over lesion type and anatomical location.
Experimental results demonstrate that incorporating lesion-level evidence consistently improves model performance across all question types, highlighting the critical role of visual grounding in enabling explainable, clinically relevant ophthalmic VQA.

%
%
%
%
\bibliographystyle{splncs04}
\bibliography{mybibliography}
\end{document}